\documentclass[runningheads]{llncs}
\usepackage[T1]{fontenc}

\usepackage{amsmath}
\usepackage{graphicx,verbatim}
\usepackage{subcaption}
\usepackage{multirow}
\usepackage{makecell}
\usepackage[table]{xcolor}

\begin{document}
\title{Imaging-Based Mortality Prediction in Patients with Systemic Sclerosis}
\author{Alec K. Peltekian\inst{1}\orcidID{0000-0001-9082-4000} \and
Karolina Senkow\inst{2}\orcidID{0000-0001-9942-3529} \and
Gorkem Durak\inst{3}\orcidID{0000-0002-1608-1955} \and
Kevin M. Grudzinski\inst{2}\orcidID{0000-0002-2317-5968} \and
Bradford C. Bemiss\inst{2}\orcidID{0000-0003-2877-4174} \and
Jane E. Dematte\inst{2}\orcidID{0000-0003-4907-9525} \and
Carrie Richardson\inst{4}\orcidID{0000-0002-2541-175X} \and
Nikolay S. Markov\inst{2}\orcidID{0000-0002-3659-4387} \and
Mary Carns\inst{2}\orcidID{0000-0002-5063-156X} \and
Kathleen Aren\inst{4}\orcidID{0000-0003-4770-2643} \and
Alexandra Soriano\inst{2}\orcidID{0009-0002-5032-3716} \and
Matthew Dapas\inst{4}\orcidID{0000-0002-8004-8535} \and
Harris Perlman\inst{4}\orcidID{0000-0003-4174-608X} \and
Aaron Gundersheimer\inst{2}\orcidID{0000-0003-0789-6412} \and
Kavitha C. Selvan\inst{2}\orcidID{0000-0002-0988-0222} \and
John Varga\inst{5}\orcidID{0000-0001-8400-687X} \and
Monique Hinchcliff\inst{6}\orcidID{0000-0002-8652-9890} \and
Krishnan Warrior\inst{2}\orcidID{0000-0002-4157-5500} \and
Catherine A Gao\inst{2}\orcidID{0000-0001-5576-3943} \and
Richard G. Wunderink\inst{2}\orcidID{0000-0002-8527-4195} \and
GR Scott Budinger\inst{2}\orcidID{0000-0002-3114-5208} \and
Alok N. Choudhary\inst{1,7}\orcidID{0000-0001-8152-6319} \and
Anthony J. Esposito\inst{2}\orcidID{0000-0002-8636-0845} \and
Alexander V Misharin\inst{2}\orcidID{0000-0003-2879-3789} \and
Ankit Agrawal\inst{7}\orcidID{0000-0002-5519-0302} \and
Ulas Bagci\inst{3}\orcidID{0000-0001-7379-6829}$^{\star}$}
\authorrunning{A. Peltekian et al.}

\institute{Department of Computer Science, Northwestern University, Evanston, IL 60208, USA \and
Division of Pulmonary and Critical Care Medicine, Northwestern University, Chicago, IL 60611, USA \and
Machine \& Hybrid Intelligence Lab, Department of Radiology, Northwestern University, Chicago, IL 60611, USA
\email{ulas.bagci@northwestern.edu} \and
Division of Rheumatology, Northwestern University, Chicago, IL 60611, USA \and
Division of Rheumatology, University of Michigan, Ann Arbor, MI 48109, USA \and
Section of Rheumatology, Allergy \& Immunology, Yale University, New Haven, CT 06520, USA \and
Department of Electrical and Computer Engineering, Northwestern University, Evanston, IL 60208, USA}
\maketitle 
\begin{abstract}
Interstitial lung disease (ILD) is a leading cause of morbidity and mortality in systemic sclerosis (SSc). Chest computed tomography (CT) is the primary imaging modality for diagnosing and monitoring lung complications in SSc patients. However, its role in disease progression and mortality prediction has not yet been fully clarified. This study introduces a novel, large-scale longitudinal chest CT analysis framework that utilizes radiomics and deep learning to predict mortality associated with lung complications of SSc. We collected and analyzed 2,125 CT scans from SSc patients enrolled in the Northwestern Scleroderma Registry, conducting mortality analyses at one, three, and five years using advanced imaging analysis techniques. Death labels were assigned based on recorded deaths over the one-, three-, and five-year intervals, confirmed by expert physicians. In our dataset, 181, 326, and 428 of the 2,125 CT scans were from patients who died within one, three, and five years, respectively. Using ResNet-18, DenseNet-121, and Swin Transformer we use pre-trained models, and fine-tuned on 2,125 images of SSc patients. Models achieved an AUC of 0.769, 0.801, 0.709 for predicting mortality within one-, three-, and five-years, respectively. Our findings highlight the potential of both radiomics and deep learning computational methods to improve early detection and risk assessment of SSc-related interstitial lung disease, marking a significant advancement in the literature.


\end{abstract}

\section{Introduction}
Systemic sclerosis (SSc) represents a significant clinical challenge with its heterogeneous presentation and unpredictable disease trajectories, characterized by progressive fibrosis affecting multiple organ systems~\cite{volkmann2016treatment}. Interstitial lung disease (ILD), the leading cause of morbidity and mortality in SSc patients, exhibits particularly variable progression patterns that have defied conventional prediction methods~\cite{hoffmann2021progressive}. Despite advances in therapeutic options, the identification of high-risk individuals who would benefit most from early intervention remains an unresolved clinical need, highlighting a critical gap between imaging capabilities and prognostic information.

While high-resolution CT remains the gold standard for ILD diagnosis, its full potential for predicting long-term mortality has been largely untapped~\cite{goldin2018longitudinal}. Traditional radiological assessments rely on qualitative interpretation, introducing observer variability and limiting the detection of subtle yet clinically significant disease patterns that may have prognostic value~\cite{goh2008interstitial}. Our innovation overcomes these limitations by implementing a systematic comparison of emerging quantitative imaging techniques---radiomics and deep learning---to extract high-dimensional imaging biomarkers that significantly improve mortality risk stratification~\cite{jacob2016automated}.
The application of transformer-based architectures to SSc-ILD mortality prediction represents a significant advancement over existing approaches. Unlike previous studies that have largely focused on idiopathic pulmonary fibrosis (IPF) with its relatively predictable trajectory, our work specifically addresses SSc-ILD's highly variable disease patterns and diverse inflammatory responses. By demonstrating that transformer-based models can capture the complex spatial relationships in fibrotic lung tissue with greater accuracy than conventional CNNs, we establish a new benchmark for imaging-based risk stratification in this challenging disease.

\textbf{Our contribution.} We present the first large-scale application of advanced artificial intelligence techniques to address this unmet clinical challenge, leveraging one of the largest annotated SSc-ILD imaging repositories from the Northwestern Scleroderma Registry. Our novel contribution lies in repurposing state-of-the-art computer vision architectures to capture disease-specific imaging biomarkers that conventional radiological assessments fail to identify. Our work is a new computer aided diagnosis application, which has not been done before.

Our comprehensive comparative analysis evaluates multiple deep learning models, including SwinUNETR~\cite{hatamizadeh2021swin}, DenseNet-121~\cite{huang2017densely}, and ResNet-18~\cite{he2016deep}, against radiomics-based machine learning approaches. This systematic evaluation across architectural paradigms provides unprecedented insight into which computational approaches best capture the imaging phenotypes associated with mortality risk in SSc-ILD. The novel integration of spatial attention mechanisms proves particularly valuable for detecting the subtle, heterogeneous patterns of fibrosis progression characteristic of this disease.

This study bridges a critical translational gap by demonstrating how advanced computer vision techniques can be effectively applied to rare disease imaging. Beyond SSc-ILD, our methodological approach establishes a template for AI-driven risk stratification in other rare fibrotic disorders with limited prior research in automated imaging-based mortality prediction. The clinical significance of our work lies in its potential to transform patient management by enabling earlier identification of high-risk individuals, facilitating personalized therapeutic decision-making, and ultimately improving long-term outcomes in this devastating disease.

\section{Methods}

\subsection{Image Data Collection}
CT imaging data were acquired from the Northwestern Scleroderma Registry, a comprehensive repository of high-resolution chest CT scans from patients with SSc. These scans were obtained as part of routine clinical care and standardized research protocols, spanning from 2001 to 2023 (Figure 1). CT scans were stored in DICOM format and preprocessed for quality assurance using ITK-SNAP. Low-resolution scans and those with insufficient slice aggregation were excluded. Data access was regulated under Northwestern Institutional Review Board (IRB) approval (STU00002669). To develop models for predicting mortality risk within one, three, and five years of a CT scan, mortality labels were assigned based on recorded death or lung transplant events, treating both as equivalent endpoints due to their clinical significance in end-stage ILD.

\begin{figure}
\centering
\includegraphics[width=0.95\textwidth,keepaspectratio]{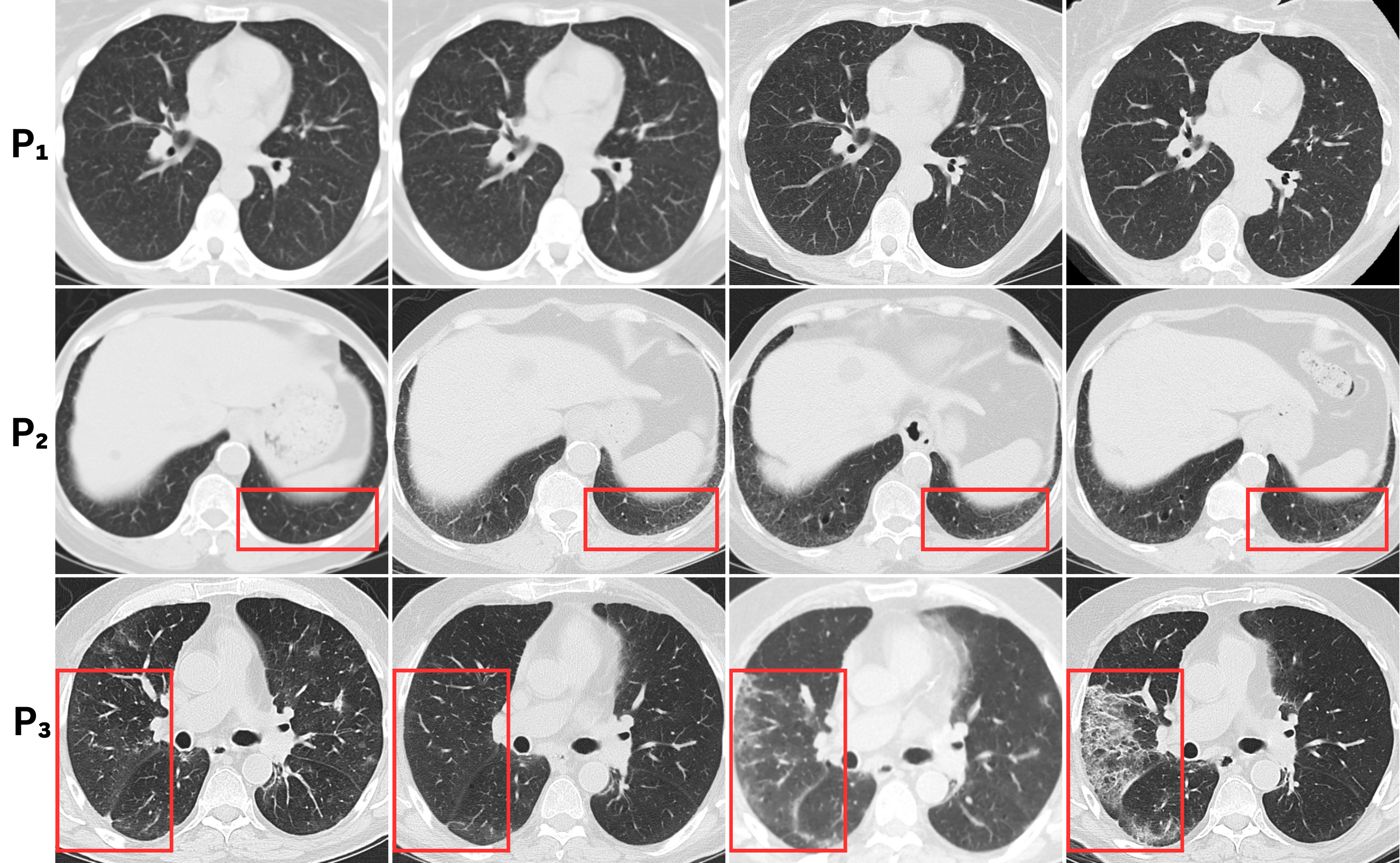}
\caption{\textbf{Longitudinal CT scans illustrating ILD progression in three patients with SSc, with red boxes highlighting disease progression}. P1 shows a patient who remained ILD-negative throughout the surveillance period. P2 shows a patient who developed ILD over time, with images taken at baseline and the 3rd, 10th, and 16th years, demonstrating disease onset and progressive fibrotic changes (note increasing reticular patterns in highlighted regions). P3 presents a patient with pre-existing ILD that rapidly worsened, with scans captured at baseline and the 1st, 2nd, and 3rd years, showing significant progression of fibrotic remodeling and honeycombing patterns within the marked areas.}
\label{fig1}
\end{figure}

\subsection{Image Preprocessing and Standardization}
We applied a structured preprocessing pipeline to the chest CT scans prior to feature extraction and model training (Figure 2). All CT scans were converted to a uniform 1mm slice thickness through isotropic resampling. This step accounted for variations in scanner types, acquisition protocols, and slice reconstruction settings, ensuring that each scan maintained a consistent voxel resolution. Prior to modeling, we performed lung segmentation using the LungMask R231 model~\cite{hofmanninger2020automatic}, a validated deep learning-based segmentation algorithm.

\subsubsection{Radiomics Processing}
In parallel, a separate radiomics-based machine learning pipeline was implemented. Instead of using full images, radiomics extracts quantitative features that describe lung shape, texture, and intensity distribution. From each pair of segmented lungs with original images, we extracted a total of 107 radiomics features using the PyRadiomics package, comprising 14 shape descriptors (e.g., elongation, flatness), 18 first-order statistics (e.g., mean intensity, entropy), and 75 texture features derived from gray-level co-occurrence and size-zone matrices.

\subsubsection{Deep Learning Segmentation of Lungs}
This segmentation model (LungMask R231) automatically identified and extracted the lung parenchyma, removing surrounding anatomical structures to focus on disease-relevant features. The segmented lung masks were then multiplied with the original CT images, retaining only the lung regions while suppressing irrelevant background information. The resulting 3D volumetric data, maintaining full spatial resolution across all three dimensions, served as input for subsequent deep learning models, enabling comprehensive analysis of fibrotic patterns throughout the entire lung volume.

\begin{figure}
\centering
\includegraphics[width=1\textwidth]{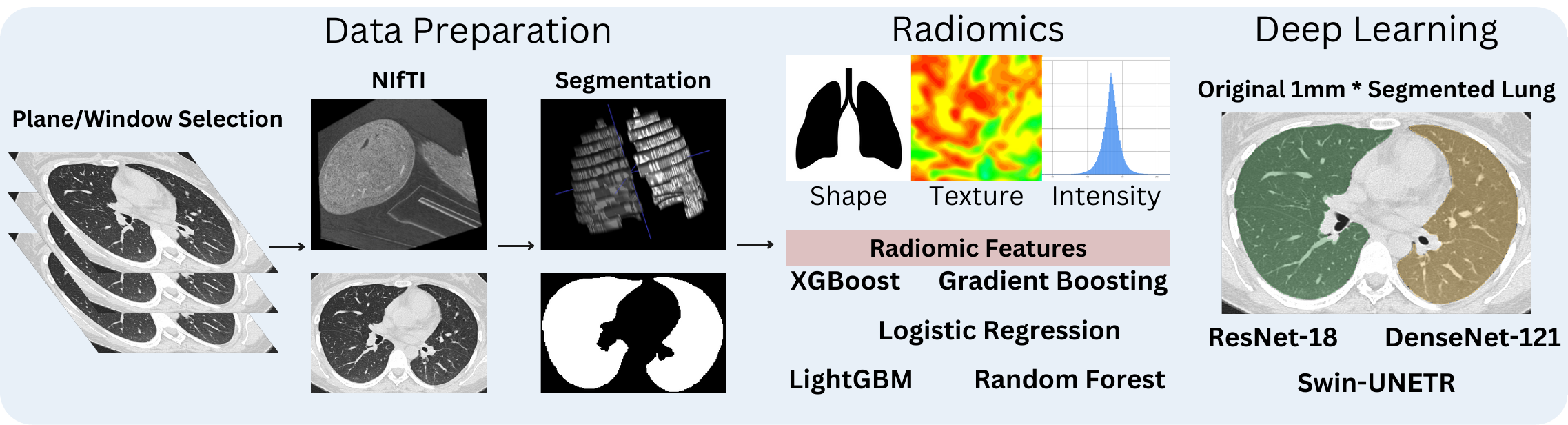}
\caption{\textbf{Overview of the data preprocessing pipeline}. \textbf{(Data Preparation)} All CT scans were resampled to a uniform 1mm slice thickness before feature extraction and segmented with LungMask R231 Model. \textbf{(Radiomics)} Radiomic features were extracted from the original/segmented lung pair, and used for training machine learning models using optuna hyperpameter optimization, selecting the best model with optimal parameters for each task. \textbf{(Deep Learning)} Segmented lung regions, multiplied with original CT images, are used as input for deep learning modeling.}
\label{fig2}
\end{figure}

\subsection{Mortality Prediction Modeling}
\sloppy

To compare traditional radiomics with DL for mortality prediction, we implemented two parallel pipelines. For radiomics, five machine learning models (XGBoost~\cite{chen2016xgboost}, Random Forest~\cite{breiman2001random}, Logistic Regression~\cite{hosmer2013applied}, LightGBM~\cite{ke2017lightgbm}, and Gradient Boosting~\cite{friedman2001greedy}) were trained on 1,284 imaging features per CT scan, including texture, shape, and intensity biomarkers. Hyperparameters (e.g., XGBoost’s tree depth, LightGBM’s bagging fraction) were optimized via Optuna~\cite{akiba2019optuna} over 500 trials, minimizing cross-entropy loss on a patient-stratified 64\%/16\%/20\% train/validation/test split to prevent data leakage. For DL, we fine-tuned Swin Transformer, ResNet-18, and DenseNet-121 on 512×512×Z lung region crops, normalized to [-1200, 600] Hounsfield Units (HU) to suppress irrelevant tissues. Swin Transformer’s shifted-window self-attention was prioritized for its ability to model multi-scale fibrosis patterns, while ResNet/DenseNet leveraged ImageNet-pretrained 2D kernels extended with 3D convolutions.

Training employed 5-fold group-stratified cross-validation to ensure patient-level separation, preventing data leakage from the same patient appearing in both training and testing sets. For each fold, GroupKFold split the data into 80\% training and 20\% holdout based on patient groups. Each fold's 20\% holdout represented a distinct patient population, ensuring that across all 5 folds, different patient groups were held out for evaluation. Within each fold's holdout, patients were further divided into validation (10\% of total data) and test sets (10\% of total data). This approach resulted in an 80\%/10\%/10\% train/validation/test split per fold, with the critical advantage that all patients (100\% of the dataset) were eventually evaluated across the 5 folds, as each patient group appeared in exactly one fold's holdout set. Training was conducted for a maximum of 50 epochs with early stopping patience of 10 epochs based on validation loss, using AdamW optimizer with learning rate of $4 \times 10^{-4}$ and weight decay of $1 \times 10^{-5}$. Models were trained with batch size of 4 due to GPU memory constraints. Weighted cross entropy loss was used to address class imbalance. Class weights were calculated using inverse frequency weighting, where each class weight was computed as the total number of samples divided by twice the class count, ensuring that minority classes received proportionally higher influence during training. Performance was evaluated via AUROC across 1/3/5-year mortality windows.

Although segmentation-focused architectures such as UNet\cite{isensee2021nnu} and nnUNet\cite{ronneberger2015u} have demonstrated strong performance for pixel-level delineation tasks, our objective centered on whole-image classification for mortality prediction. We therefore selected the Swin Transformer for its hierarchical attention mechanism, which is particularly effective at capturing spatial dependencies and the complex fibrosis patterns characteristic of volumetric chest CT in SSc-ILD.

\section{Results}
\subsection{Imaging Data Overview}
We analyzed a total of 2,125 CT scans from the Northwestern Scleroderma Registry collected between 2001 and 2023. The distribution of CT scans over time (Figure 3A) shows a steady increase in imaging utilization, likely reflecting both improved disease surveillance strategies and evolving clinical guidelines emphasizing early ILD detection. While some patients had only a single baseline CT scan, others had multiple follow-ups, reflecting different disease trajectories and monitoring intensities.

\begin{figure}
\centering
\includegraphics[width=1\textwidth]{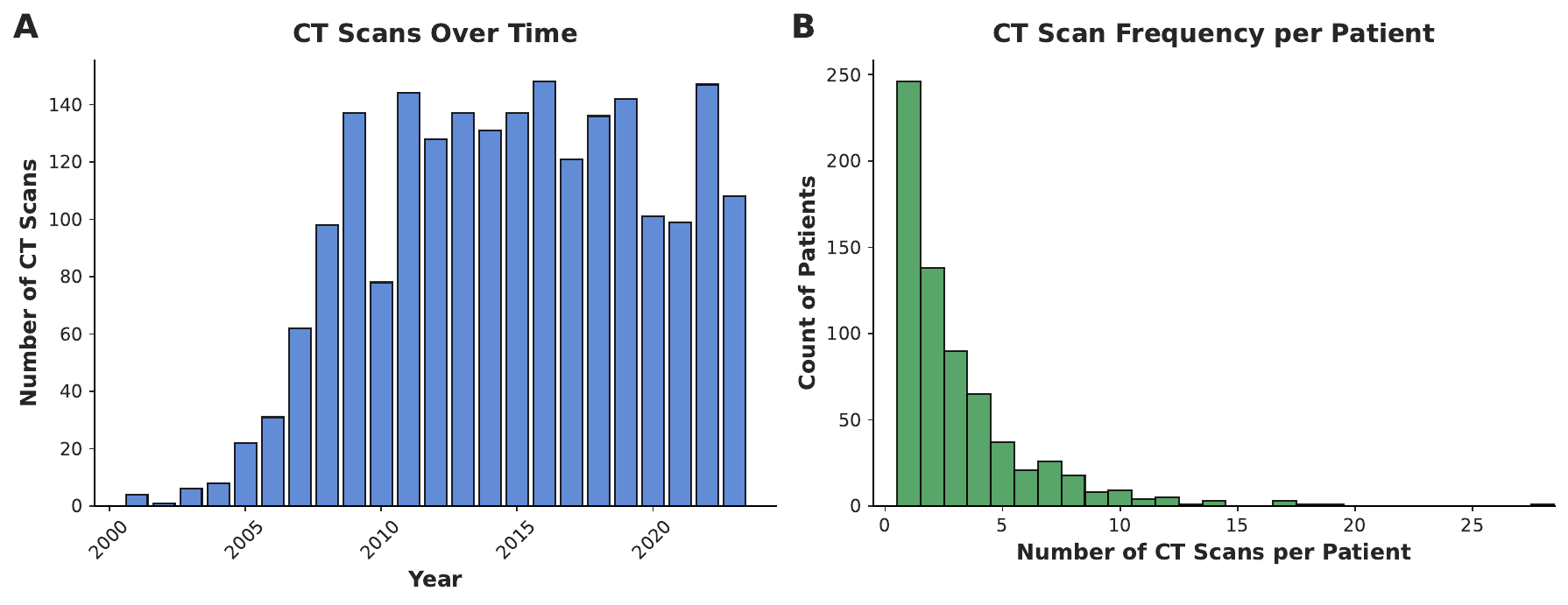}
\caption{ \textbf{CT scan utilization and ILD survival patterns.} 
\textbf{(A)} Annual distribution of CT scans from the Northwestern Scleroderma Registry (2001–2023), illustrating an increasing trend in CT imaging utilization. 
\textbf{(B)} Distribution of survival times from the first ILD-confirmed CT scan to death, demonstrating significant variability in survival outcomes.}
\label{fig3}
\end{figure}

Given the central role of ILD in SSc mortality, we analyzed survival patterns with CT scans (Figure 3B). There is great variability in disease progression, as some patients survived for over a decade following ILD detection, while others experienced rapid deterioration within a few years. This heterogeneity highlights the urgent need for improved risk stratification strategies to differentiate patients with high risk of mortality from those with more stable disease courses. By integrating longitudinal CT data into predictive models, we aim to address this variability and develop more precise, personalized risk assessment tools.

\subsection{Survival Trends and Mortality Risk}
Given the established link between ILD and SSc-related mortality, we conducted a survival analysis to assess overall trends in the cohort. A significant proportion of patients survived more than 10 years after their initial ILD-positive CT, though many faced accelerated disease progression within shorter time frames (Figure 4A). Most patients received CT scans within a defined window before death, suggesting that late-stage imaging patterns may have distinct prognostic implications (Figure 4B).
These findings suggest that CT imaging, particularly when analyzed in relation to mortality timing, may serve as a valuable biomarker for disease progression and prognosis. By integrating longitudinal CT data into predictive modeling, we can improve our ability to identify high-risk patients earlier and develop more targeted monitoring and intervention strategies.

\begin{figure}
\centering
\includegraphics[width=1\textwidth]{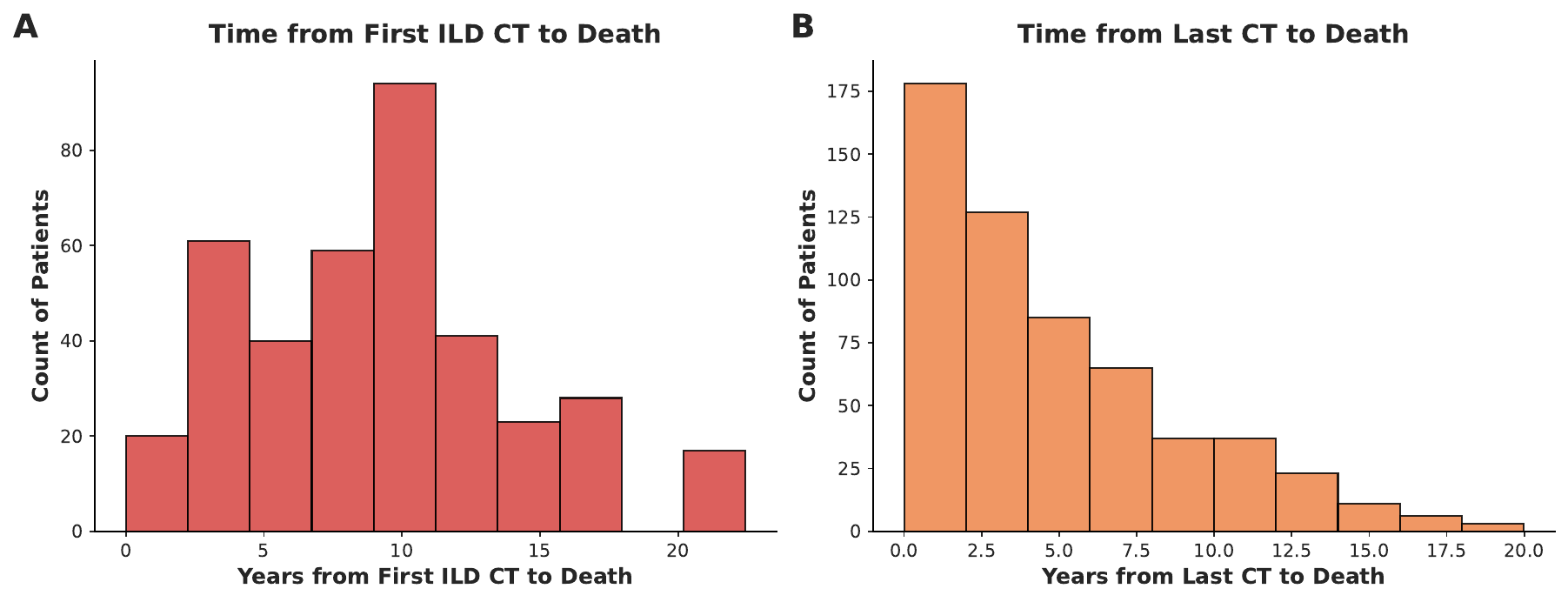}
\caption{\textbf{Mortality timing relative to CT scans.} 
\textbf{(A)} Time from a patient’s first ILD-confirmed CT scan to death, emphasizing variability in long-term survival after initial ILD detection. 
\textbf{(B)} Time from the last recorded CT scan to death, showing that most patients undergo imaging within a specific window before mortality}
\label{fig4}
\end{figure}

\subsection{Predictive Modeling Results}
We evaluated the performance of radiomics-based machine learning models and deep learning architectures in predicting mortality risk at 1-, 3-, and 5-year intervals. Models were assessed using area under the receiver operating characteristic curve (AUROC), sensitivity, specificity, precision, and F1-score. A comparative summary of performance across all models is presented in Table~\ref{table:model_results}.

\renewcommand{\arraystretch}{1.3}

\begin{table}[h]
    \centering
    \begin{tabular}{|c|c|c|c|c|c|c|}
        \hline
        \rowcolor{gray!40}
        \textbf{Task} & \textbf{Model} & \textbf{AUROC} & \textbf{Sensitivity} & \textbf{Specificity} & \textbf{Precision} & \textbf{F1-Score} \\
        \hline
        \cellcolor{gray!15} \multirow{4}{*}{\makecell{\textbf{1-Year} \\ \textbf{Mortality}}} 
        & Logist-Reg & 0.723 & \textbf{0.500} & 0.751 & 0.070 & 0.125 \\
        & \cellcolor{gray!15} Swin UNETR  & \textbf{0.769} & 0.168 & 0.975 & \textbf{0.342} & 0.184 \\
        & DenseNet-121 & 0.765 & 0.261 & 0.972 & 0.323 & \textbf{0.269} \\
        & \cellcolor{gray!15} ResNet-18 & 0.750 & 0.146 & \textbf{0.981} & 0.104 & 0.104 \\
        \hline
        \cellcolor{gray!15} \multirow{4}{*}{\makecell{\textbf{3-Year} \\ \textbf{Mortality}}} 
        & RandForest & 0.682 & \textbf{0.640} & 0.635 & 0.186 & 0.288 \\
        & \cellcolor{gray!15} Swin UNETR & \textbf{0.801} & 0.440 & 0.967 & \textbf{0.647} & \textbf{0.524} \\
        & DenseNet-121 & 0.703 & 0.249 & \textbf{0.972} & 0.518 & 0.323 \\
        & \cellcolor{gray!15} ResNet-18 & 0.711 & 0.195 & 0.946 & 0.349 & 0.224 \\
        \hline
        \cellcolor{gray!15} \multirow{4}{*}{\makecell{\textbf{5-Year} \\ \textbf{Mortality}}} 
        & Logist-Reg & 0.707 & \textbf{0.577} & 0.761 & 0.346 & \textbf{0.433} \\
        & \cellcolor{gray!15} Swin UNETR & 0.609 & 0.075 & \textbf{0.968} & 0.117 & 0.083 \\
        & DenseNet-121 & \textbf{0.709} & 0.237 & 0.930 & \textbf{0.390} & 0.279 \\
        & \cellcolor{gray!15} ResNet-18 & 0.688 & 0.188 & 0.919 & 0.151 & 0.149 \\
        \hline
    \end{tabular}
    \caption{Performance comparison of radiomics-based and deep learning models for predicting 1-, 3-, and 5-year mortality in SSc patients.}
    \label{table:model_results}
\end{table}

There was a significant class imbalance in our mortality labels, with a relatively small proportion of CT scans belonging to patients who died within the specified time frames. Out of a total of 2,125 CT scans, only 181 (8.5\%) corresponded to deaths within one year, 326 (15.3\%) within three years, and 428 (20.1\%) within five years.

Radiomics-based models, including Logistic Regression (LR) and Random Forest (RF), provided predictions based on extracted imaging features. However, their performance was moderate across all prediction windows, with AUROCs ranging from 0.682 to 0.723. Despite this, radiomics models demonstrated higher specificity compared to deep learning models, making them useful for confirming low-risk cases.

Deep learning models, trained directly on segmented CT images, outperformed radiomics models across most tasks. Swin UNETR, a transformer-based model, achieved the highest AUROC (0.769) for 1-year and 3-year mortality prediction (AUROC = 0.801), likely due to its ability to model long-range dependencies and complex feature interactions. This suggests that transformer-based architectures may be particularly well-suited for mid-term mortality risk assessment in SSc-ILD. DenseNet had the best performance for the 5-year mortality prediction (AUROC = 0.709). As the mortality window prediction extends, the task becomes more difficult.

Deep learning methods can capture subtle disease-related features, whereas radiomics models rely on predefined feature extraction, making them easier to integrate with traditional clinical workflows. Overall, these results suggest that combining radiomics-based and deep learning models could improve the prediction of mortality risk. Future work should focus on multimodal approaches that integrate imaging biomarkers with functional and clinical data.

\section{Discussion and Conclusion}
Our study demonstrates that advanced deep learning models, particularly transformer-based architectures, significantly outperform traditional approaches in extracting prognostic imaging biomarkers from CT scans of systemic sclerosis patients. The Swin Transformer's hierarchical attention mechanism excels at capturing complex spatial relationships and subtle patterns associated with disease progression, providing superior mortality risk prediction compared to conventional CNN and radiomics methods.
The technical innovation of our approach enables early identification of high-risk patients through attention-guided feature extraction, potentially creating a critical window for therapeutic intervention. The observed heterogeneity in survival trajectories underscores the clinical importance of accurate risk stratification tools.

While our current approach achieved strong performance using standard preprocessing techniques, future implementations could benefit from data augmentation strategies, multi-modal transformer architectures that integrate imaging with clinical data, and incorporation of uncertainty quantification. Careful consideration must be given to preserving the clinical realism of fibrotic patterns when applying transformations to medical imaging data. This research establishes a new benchmark for AI-driven mortality prediction in systemic sclerosis and provides a foundation for clinical decision support systems that could significantly improve patient outcomes through personalized treatment strategies.

\section*{Acknowledgments}
This research was supported in part through a generous gift from K. Querrey and L. Simpson. This research was also supported by the computational resources and staff contributions provided for the Quest high-performance computing facility at Northwestern University, which is jointly supported by the Office of the Provost, the Office for Research, and Northwestern University Information Technology. This research was also supported in part through the computational resources and staff contributions provided by the Genomics Compute Cluster, which is jointly supported by the Feinberg School of Medicine, the Center for Genetic Medicine, Feinberg's Department of Biochemistry and Molecular Genetics, the Office of the Provost, the Office for Research, and Northwestern Information Technology. The Genomics Compute Cluster is part of Quest, Northwestern University's high-performance computing facility, with the purpose of advancing research in genomics. M.H. was supported by the NIH (grant no. R01AR07327). C.A.G. was supported by the NIH (grant no. K23HL169815), a Parker B. Francis Opportunity Award, and an American Thoracic Society Unrestricted Grant. R.G.W. is supported by the NIH (grant nos. U19AI135964, R01AI158530, R01HL149883, P01HL154998, U01TR003528). G.R.S.B. was supported by a Chicago Biomedical Consortium grant, Northwestern University Dixon Translational Science Award, Simpson Querrey Lung Institute for Translational Science, the NIH (grant nos. P01AG049665, P01HL154998, U54AG079754, R01HL147575, R01HL158139, R01HL147290, R21AG075423 and U19AI135964), and the Veterans Administration (award no. I01CX001777). A.V.M. was supported by the NIH (grant nos. U19AI135964, P01AG049665, P01HL154998, U19AI181102, R01HL153312, R01HL158139, R01ES034350 and R21AG075423). A.A. was supported by the NIH (grant nos. U19AI135964 and R01HL158138) and Simpson Querrey Lung Institute for Translational Science.  A.J.E. was supported by the NIH (grant no. L30HL149048). U.B. acknowledges the following grant: R01-HL171376. The funders had no role in the study design, data collection and analysis, decision to publish, or preparation of the manuscript.

\bibliographystyle{splncs04}
\bibliography{references}

\begin{thebibliography}{10}
\providecommand{\url}[1]{\texttt{#1}}
\providecommand{\urlprefix}{URL }
\providecommand{\doi}[1]{https://doi.org/#1}

\bibitem{akiba2019optuna}
Akiba, T., Sano, S., Yanase, T., Ohta, T., Koyama, M.: Optuna: A next-generation hyperparameter optimization framework. In: Proceedings of the 25th ACM SIGKDD international conference on knowledge discovery \& data mining. pp. 2623--2631 (2019)

\bibitem{breiman2001random}
Breiman, L.: Random forests. Machine learning  \textbf{45},  5--32 (2001)

\bibitem{chen2016xgboost}
Chen, T., Guestrin, C.: Xgboost: A scalable tree boosting system. In: Proceedings of the 22nd acm sigkdd international conference on knowledge discovery and data mining. pp. 785--794 (2016)

\bibitem{friedman2001greedy}
Friedman, J.H.: Greedy function approximation: a gradient boosting machine. Annals of statistics pp. 1189--1232 (2001)

\bibitem{goh2008interstitial}
Goh, N.S., Desai, S.R., Veeraraghavan, S., Hansell, D.M., Copley, S.J., Maher, T.M., Corte, T.J., Sander, C.R., Ratoff, J., Devaraj, A., et~al.: Interstitial lung disease in systemic sclerosis: a simple staging system. American journal of respiratory and critical care medicine  \textbf{177}(11),  1248--1254 (2008)

\bibitem{goldin2018longitudinal}
Goldin, J.G., Kim, G.H.J., Tseng, C.H., Volkmann, E., Furst, D., Clements, P., Brown, M., Roth, M., Khanna, D., Tashkin, D.P.: Longitudinal changes in quantitative interstitial lung disease on ct after immunosuppression in the scleroderma lung study ii. Annals of the American Thoracic Society  \textbf{15}(11),  1286--1295 (2018)

\bibitem{hatamizadeh2021swin}
Hatamizadeh, A., Nath, V., Tang, Y., Yang, D., Roth, H.R., Xu, D.: Swin unetr: Swin transformers for semantic segmentation of brain tumors in mri images. In: International MICCAI brainlesion workshop. pp. 272--284. Springer (2021)

\bibitem{he2016deep}
He, K., Zhang, X., Ren, S., Sun, J.: Deep residual learning for image recognition. In: Proceedings of the IEEE conference on computer vision and pattern recognition. pp. 770--778 (2016)

\bibitem{hoffmann2021progressive}
Hoffmann-Vold, A.M., Allanore, Y., Alves, M., Brunborg, C., Air{\'o}, P., Ananieva, L.P., Czirj{\'a}k, L., Guiducci, S., Hachulla, E., Li, M., et~al.: Progressive interstitial lung disease in patients with systemic sclerosis-associated interstitial lung disease in the eustar database. Annals of the rheumatic diseases  \textbf{80}(2),  219--227 (2021)

\bibitem{hofmanninger2020automatic}
Hofmanninger, J., Prayer, F., Pan, J., R{\"o}hrich, S., Prosch, H., Langs, G.: Automatic lung segmentation in routine imaging is primarily a data diversity problem, not a methodology problem. European radiology experimental  \textbf{4},  1--13 (2020)

\bibitem{hosmer2013applied}
Hosmer~Jr, D.W., Lemeshow, S., Sturdivant, R.X.: Applied logistic regression. John Wiley \& Sons (2013)

\bibitem{huang2017densely}
Huang, G., Liu, Z., Van Der~Maaten, L., Weinberger, K.Q.: Densely connected convolutional networks. In: Proceedings of the IEEE conference on computer vision and pattern recognition. pp. 4700--4708 (2017)

\bibitem{isensee2021nnu}
Isensee, F., Jaeger, P.F., Kohl, S.A., Petersen, J., Maier-Hein, K.H.: nnu-net: a self-configuring method for deep learning-based biomedical image segmentation. Nature methods  \textbf{18}(2),  203--211 (2021)

\bibitem{jacob2016automated}
Jacob, J., Bartholmai, B., Rajagopalan, S., Kokosi, M., Nair, A., Karwoski, R., Raghunath, S., Walsh, S., Wells, A., Hansell, D.: Automated quantitative ct versus visual ct scoring in idiopathic pulmonary fibrosis: validation against pulmonary function. J Thorac Imaging  \textbf{31}(304), ~11 (2016)

\bibitem{ke2017lightgbm}
Ke, G., Meng, Q., Finley, T., Wang, T., Chen, W., Ma, W., Ye, Q., Liu, T.Y.: Lightgbm: A highly efficient gradient boosting decision tree. Advances in neural information processing systems  \textbf{30} (2017)

\bibitem{ronneberger2015u}
Ronneberger, O., Fischer, P., Brox, T.: U-net: Convolutional networks for biomedical image segmentation. In: Medical image computing and computer-assisted intervention--MICCAI 2015: 18th international conference, Munich, Germany, October 5-9, 2015, proceedings, part III 18. pp. 234--241. Springer (2015)

\bibitem{volkmann2016treatment}
Volkmann, E.R., Tashkin, D.P.: Treatment of systemic sclerosis--related interstitial lung disease: a review of existing and emerging therapies. Annals of the American Thoracic Society  \textbf{13}(11),  2045--2056 (2016)

\end{thebibliography}
\end{document}